\documentclass[letterpaper]{article}

\usepackage{aaai21}  % DO NOT CHANGE THIS
\usepackage{times}  % DO NOT CHANGE THIS
\usepackage{helvet} % DO NOT CHANGE THIS
\usepackage{courier}  % DO NOT CHANGE THIS
\usepackage[hyphens]{url}  % DO NOT CHANGE THIS
\usepackage{graphicx} % DO NOT CHANGE THIS
\usepackage{natbib}  % DO NOT CHANGE THIS AND DO NOT ADD ANY OPTIONS TO IT
\usepackage{caption} % DO NOT CHANGE THIS AND DO NOT ADD ANY OPTIONS TO IT
\usepackage{helvet}
\usepackage{courier}
\usepackage{amssymb} %maths
\usepackage{amsmath} %maths
\usepackage[ruled,vlined,linesnumbered]{algorithm2e}
\usepackage{dsfont}
\usepackage{booktabs} % nicer tables :-)
\usepackage{multirow}   % afor multi-row tables
\usepackage{array, makecell} %
\usepackage{graphics}
\usepackage{amsthm}
\usepackage{balance}

\makeatletter
\newcommand{\pushright}[1]{\ifmeasuring@#1\else\omit\hfill$\equationstyle#1$\fi\ignorespaces}
\newcommand{\pushleft}[1]{\ifmeasuring@#1\else\omit$\equationstyle#1$\hfill\fi\ignorespaces}
\makeatother

\DeclareMathOperator*{\esssup}{ess\,sup}

\usepackage{multicol}
\usepackage[noend]{algorithmic}
\usepackage{algorithm2e}
\usepackage[english]{babel}
\usepackage{subfigure}
\usepackage{mathtools}
\mathtoolsset{showonlyrefs=false}
\makeatletter

 \DeclareMathOperator*{\argmin}{argmin}
\newtheorem{theorem}{\protect\theoremname}
\newtheorem{defn}{\protect\definitionname}
\newtheorem{proposition}{\protect\propositionname}

\newtheorem{assum}{\protect\assumname}
\newtheorem{problem}{\protect\probname}
\newtheorem{corollary}{\protect\corname}

\providecommand{\definitionname}{\textbf{Definition}}
\providecommand{\propositionname}{\textbf{Proposition}}
\providecommand{\remarkname}{\textbf{Remark}}
\providecommand{\theoremname}{\textbf{Theorem}}
\providecommand{\lemmaname}{Lemma}
\providecommand{\assumname}{\textbf{Assumption}}
\providecommand{\probname}{\textbf{Problem}}
\providecommand{\corname}{\textbf{Corollary}}

\title{Constrained Risk-Averse Markov Decision Processes}
\author {
Mohamadreza Ahmadi$^{1}$, Ugo Rosolia$^{1}$, \\ Michel D. Ingham$^{2}$, Richard M. Murray$^{1}$, and Aaron D. Ames$^{1}$ \\  }
\affiliations{
     $^{1}$California Institute of Technology, 1200 E. California Blvd., Pasadena, CA 91125\\
$^{2}$NASA Jet Propulsion Laboratory, 4800 Oak Grove Dr, Pasadena, CA 91109.\\ 
    {mrahmadi@caltech.edu}
}
\begin{document}
\maketitle
\begin{abstract}
\begin{quote}
We consider the problem of designing policies for Markov decision processes (MDPs) with dynamic coherent risk objectives and constraints. We begin by formulating the problem in a Lagrangian framework. Under the assumption that the risk objectives and constraints can be represented by a Markov risk transition mapping, we propose an optimization-based method to synthesize Markovian policies that lower-bound the constrained risk-averse problem.  We demonstrate that the formulated optimization problems are in the form of difference convex programs (DCPs) and can be solved by the disciplined convex-concave programming (DCCP) framework. We show that these results generalize linear programs for constrained MDPs with total discounted expected costs and constraints. Finally, we illustrate the effectiveness of the proposed method with numerical experiments on a rover navigation problem involving conditional-value-at-risk (CVaR) and entropic-value-at-risk (EVaR) coherent risk measures.
\end{quote}
\end{abstract}

\section{Introduction}

\noindent With the rise of autonomous systems being deployed in real-world settings, the associated risk that stems from unknown and unforeseen circumstances is correspondingly on the rise. In particular, in risk-sensitive scenarios, such as aerospace applications, decision making should account for uncertainty and minimize the impact of unfortunate events. For example, spacecraft control technology relies heavily on a relatively large and highly skilled mission operations
team that generates detailed time-ordered and event-driven sequences of commands. This approach will
not be viable in the future with increasing number of missions and a desire to limit the operations team
and Deep Space Network (DSN) costs.
In order to maximize the science returns under these conditions, the ability to deal with emergencies and
safely explore remote regions are becoming increasingly important~\cite{mcghan2016resilient}. For instance, in Mars rover navigation problems, finding planning policies that minimize risk is critical to mission success,  due to the uncertainties present in Mars surface data~\cite{ono2018mars} (see  Figure~1).

\begin{figure}[t] \centering{
\includegraphics[scale=.3]{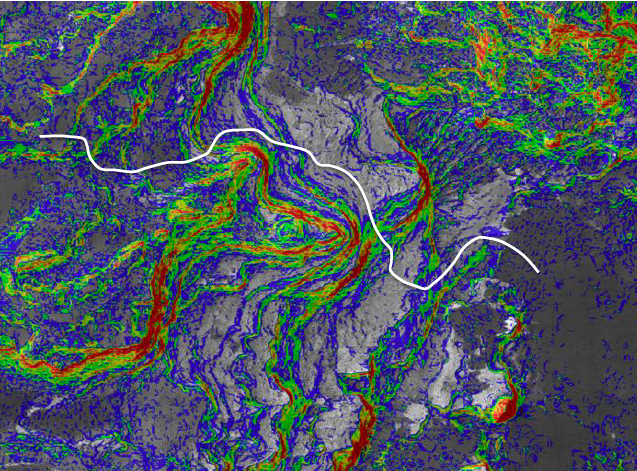}
%\vspace{-1cm}
\caption{Mars surface (Eberswalde Crater) slope uncertainty for rover navigation: the slopes range within (blue) $5^\circ-10^\circ$, (green) $10^\circ-15^\circ$, (yellow) $15^\circ-20^\circ$, (orange) $20^\circ-25^\circ$, (red)  $\ge 25^\circ$, and (the rest) $<5^\circ$ or no data. 
}}
\vspace{-0.4cm}
\label{fig:marssurface}
 \end{figure}

Risk can be quantified in numerous ways, such as chance constraints~\cite{ono2015chance,wang2020non}. However, applications in autonomy and robotics require more ``nuanced assessments of risk''~\cite{majumdar2020should}. Artzner \textit{et. al.}~\cite{artzner1999coherent} characterized a set of natural properties that are desirable for a risk measure, called a coherent risk measure, and  have  obtained widespread
acceptance in finance and operations research, among other fields. An important example of a coherent risk measure is the conditional value-at-risk (CVaR) that has received significant attention in decision making problems, such as Markov decision processes (MDPs)~\cite{chow2015risk,chow2014algorithms,prashanth2014policy,bauerle2011markov}. 

% Entropic value-at-risk (EVaR) is more recently proposed coherent risk measure, which has superior computational and mathematical properties~\cite{ahmadi2012entropic}.

% Finding policies that minimize total coherent risk measures for MDPs was studied in~\cite{ruszczynski2010risk}, wherein it was further assumed the risk measure is \emph{time consistent}, similar to the dynamic programming property. Following the footsteps of the latter contribution, ~\cite{tamar2016sequential} proposed a sampling-based algorithm for MDPs with total coherent risk measure costs using policy gradient and actor-critic methods (also, see a model predictive control technique for linear dynamical systems with coherent risk objectives~\cite{singh2018framework}). Recently, a bounded policy iteration method for designing stochastic finite state controllers for partially observable Markov decision processes (POMDPs) was proposed in~\cite{ahmadi2020risk}.

In many real world applications, \textit{risk-averse} decision making should account for system constraints, such as fuel budget, communication rates, and etc. In this paper, we attempt to address this issue and therefore consider MDPs with both total coherent risk costs and constraints. Using a Lagrangian framework and properties of coherent risk measures, we propose an optimization problem, whose solution provides a lower bound to the constrained risk-averse MDP problem. We show that this result is indeed a generalization of constrained MDPs with total expected cost costs and constraints~\cite{altman1999constrained}. For general coherent risk measures, we show that this optimization problem is a difference convex program (DCP) and propose a method based on disciplined convex-concave programming to solve it.  We illustrate our proposed method with a numerical example of path planning under uncertainty with not only CVaR risk measure but also the more recently proposed entropic value-at-risk (EVaR)~\cite{ahmadi2012entropic,mcallisteragbots} coherent risk measure.

\textbf{Notation: } We denote by $\mathbb{R}^n$ the $n$-dimensional Euclidean space and $\mathbb{N}_{\ge0}$ the set of non-negative integers. Throughout the paper, we use bold font to denote a vector and $(\cdot)^\top$ for its transpose, \textit{e.g.,} $\boldsymbol{a}=(a_1,\ldots,a_n)^\top$, with $n\in \{1,2,\ldots\}$. For a vector $\boldsymbol{a}$, we use $\boldsymbol{a}\succeq (\preceq) \boldsymbol{0}$ to denote element-wise non-negativity (non-positivity) and $\boldsymbol{a}\equiv \boldsymbol{0}$ to show all elements of $\boldsymbol{a}$ are zero. For two vectors $a,b \in \mathbb{R}^n$, we denote their inner product by $\langle \boldsymbol{a}, \boldsymbol{b} \rangle$, \textit{i.e.,} $\langle \boldsymbol{a}, \boldsymbol{b} \rangle=\boldsymbol{a}^\top \boldsymbol{b}$. For a finite set $\mathcal{A}$, we denote its power set by $2^\mathcal{A}$, \textit{i.e.,} the set of all subsets of $\mathcal{A}$. For  a probability space $(\Omega, \mathcal{F}, \mathbb{P})$ and a constant $p \in [1,\infty)$, $\mathcal{L}_p(\Omega, \mathcal{F}, \mathbb{P})$ denotes the vector space of real valued random variables $c$ for which $\mathbb{E}|c|^p < \infty$.

% For a vector $\alpha \in \mathbb{R}^n$ and an integer $s \in \{1,\ldots, n\}$ we use $\alpha(s)$ to denote the $s$th component of the vector $\alpha$ and $\alpha^\top$ to indicate the transpose. For a function $V : \mathbb{R}^n \rightarrow \mathbb{R}$, $V(\alpha)$ denotes the value of the function $V$ at $\alpha \in\mathbb{R}^n$. Throughout the paper we will use capital letters to indicate functions and lower letters to indicate vectors.

\section{Preliminaries}

In this section, we briefly review some notions and definitions used throughout the paper.

\subsection{Markov Decision Processes} \label{sec:POMDP}

%\begin{defn}[MDP] \label{defn:MDP} {
An \emph{MDP} is a tuple $\mathcal{M}=(\mathcal{S},Act, T, \kappa_0)$ consisting of a set of states $\mathcal{S}  =
\{s_{1} ,\dots,s_{|\mathcal{S}|} \}$ of the
autonomous agent(s) and world model, actions $Act = \{\alpha_{1},\dots,\alpha_{|Act|}\}$ available to the robot,
% \item Observations $\mathcal{O} = \{o_{1},\dots,o_{|\mathcal{O}|}\}$,
%\item Atomic propositions $AP = \{p_1,p_2,\dots p_{|AP|}\}$,
a transition function $T(s_{j} |s_{i} ,\alpha)$, and $\kappa_0$ describing the initial distribution over the states.

%\end{defn}
%\vspace{0.2cm}

% An MDP where the states $s\in \mathcal{S}$ are not directly observable is called a POMDP.
% A \emph{POMDP}, $\mathcal{PM}$, consists of all the elements of an MDP and in addition has a set of observations $\mathcal{O} = \{o_{1},\dots,o_{|\mathcal{O}|}\}$, and
% an observation model $O(o\mid s)$.
%\item Atomic propositions $AP = \{p_1,p_2,\dots p_{|AP|}\}$,
% \item A Transition function $T(s_{j} |s_{i} ,\alpha)$,
% \item A cost, $c(s_{i},\alpha_i ) \ge 0$, for each state $s_{i}  \in \mathcal{S} $ and action~$\alpha_i \in {Act}$.

%
This paper considers {\em finite} Markov decision processes, where $\mathcal{S} $, and $Act$
are finite sets. For each action the probability of making a transition from state $s_{i}  \in
\mathcal{S} $ to state $s_{j}  \in \mathcal{S} $ under action $\alpha \in Act$ is given by
$T(s_{j} |s_{i} ,\alpha)$. 
% In a POMDP, for each state $s_{i} $, an observation $o \in
% \mathcal{O}$ is generated independently with probability $O(o|s_{i} )$. The starting world
% state is given by the distribution $\kappa_0(s_i )$. 

The probabilistic components of a
Markov decision process must satisfy the following:
\begin{equation*}
    \begin{cases}
    \sum_{s  \in \mathcal{S} } T(s |s_{i} ,\alpha) = 1, & \forall s_i  \in \mathcal{S} ,\alpha \in Act, \\
    \sum_{s  \in \mathcal{S} } \kappa_0(s ) = 1. & {}
    \end{cases}
\end{equation*}

\subsection{Coherent Risk Measures}

Consider a probability space $(\Omega, \mathcal{F}, \mathbb{P})$, a filteration $\mathcal{F}_0 \subset \cdots \mathcal{F}_N \subset \mathcal{F} $, and an adapted sequence of random variables~(stage-wise costs) $c_t,~t=0,\ldots, N$, where $N \in \mathbb{N}_{\ge 0} \cup \{\infty\}$.
%We posit that $\mathcal{F}_0=\{\Omega, \emptyset\}$, \textit{i.e.}, $Z_1$ is deterministic.
For $t=0,\ldots,N$, we further define the spaces $\mathcal{C}_t = \mathcal{L}_p(\Omega, \mathcal{F}_t, \mathbb{P})$, $p \in [0,\infty)$,  $\mathcal{C}_{t:N}=\mathcal{C}_t\times \cdots \times \mathcal{C}_N$ and $\mathcal{C}=\mathcal{C}_0\times \mathcal{C}_1 \times \cdots$. We  assume that the sequence $\boldsymbol{c} \in \mathcal{C}$ is almost surely bounded (with exceptions having probability zero), \textit{i.e.}, 
 $
\max_t \esssup~| c_t(\omega) | < \infty.
$

In order to describe how one can evaluate the risk of sub-sequence $c_t,\ldots, c_N$ from the perspective of stage $t$, we require the following definitions.

% Consider a probability space $(\Omega, \mathcal{F}, P)$, a filteration $\{ \mathcal{F}_t \}$ on  $(\Omega, \mathcal{F}, P)$, and an adapted sequence of random variables~(stage-wise costs) $c_t,~t=1,2,\ldots$.
% %We posit that $\mathcal{F}_0=\{\Omega, \emptyset\}$, \textit{i.e.}, $Z_1$ is deterministic.
% We further define the spaces $\mathcal{C}_t = \mathcal{L}_p(\Omega, \mathcal{F}_t, P)$, $p \in [0,\infty)$, $t=1,2,\ldots$ and let $\mathcal{C}=\mathcal{C}_1\times \mathcal{C}_2 \times \cdots$. We further assume that the sequence $Z \in \mathcal{C}$ is almost surely bounded, \textit{i.e.}, 
% $$
% \max_t \mathrm{essup}~| c_t(\omega) | < \infty.
% $$
\vspace{0.2cm}
\begin{defn}[Conditional Risk Measure]{
A mapping $\rho_{t:N}: \mathcal{C}_{t:N} \to \mathcal{C}_{t}$, where $0\le t\le N$, is called a \emph{conditional risk measure}, if it has the following monoticity property:
\begin{equation*}
    \rho_{t:N}(\boldsymbol{c}) \le   \rho_{t:N}(\boldsymbol{c}'), \quad \forall \boldsymbol{c}, \forall \boldsymbol{c}' \in \mathcal{C}_{t:N}~\text{such that}~\boldsymbol{c} \preceq \boldsymbol{c}'.
\end{equation*}
}
\end{defn}
\vspace{0.2cm}
\begin{defn}[Dynamic Risk Measure]
{A \emph{dynamic risk measure} is a sequence of conditional risk measures $\rho_{t:N}:\mathcal{C}_{t:N}\to \mathcal{C}_{t}$, $t=0,\ldots,N$.}
\end{defn}
\vspace{0.2cm}
One fundamental property of dynamic risk measures is their consistency over time~\cite[Definition 3]{ruszczynski2010risk}. That is, if $c$ will be as good as $c'$ from the perspective of some future time $\theta$, and they are identical between time $\tau$ and $\theta$, then $c$ should not be worse than $c'$ from the perspective at time $\tau$.
% \vspace{0.2cm}
% \begin{defn}[Time-Consistent Risk Measure]{
% A dynamic risk measure $\left\{ \rho_{t:N}  \right\}_{t=0}^T$ is called \emph{time-consistent} if for all $0\le t \le \tau < \theta  \le T$ and all sequences $Z,W \in \mathcal{C}_{t:N}$ the conditions
% \begin{multline*}
% c_t =c'_t,~t=\tau,\ldots,\theta-1,~~\text{and}~\\ \rho_{\theta,T}(Z_\theta,\ldots,c_t) \le  \rho_{\theta,T}(W_\theta,\ldots,c'_t)
% \end{multline*}
% imply
% \begin{equation}
% \rho_{\tau,N}(c_\tau,\ldots,c_t) \le  \rho_{\tau,N}(c'_\tau,\ldots,c'_t).
% \end{equation}
% }
% \end{defn}
% \vspace{0.2cm}
If a risk measure is time-consistent, we can define the one-step conditional risk measure $\rho_t:\mathcal{C}_{t+1}\to \mathcal{C}_t$, $t=0,\ldots,N-1$ as follows:
\begin{equation}
    \rho_t(c_{t+1}) = \rho_{t,t+1}(0,c_{t+1}),
\end{equation}
and for all $t=1,\ldots,N$, we obtain:
\begin{multline}
    \label{eq:dynriskmeasure}
    \rho_{t,N}(c_t,\ldots,c_N)= \rho_t \big(c_t + \rho_{t+1} ( c_{t+1}+\rho_{t+2}(c_{t+2}+\cdots\\
    +\rho_{N-1}\left(c_{N-1}+\rho_{N}(c_N) \right) \cdots )) \big).
\end{multline}
Note that the time-consistent risk measure is completely defined by one-step conditional risk measures $\rho_t$, $t=0,\ldots,N-1$ and, in particular, for $t=0$, \eqref{eq:dynriskmeasure} defines a risk measure of the entire sequence $\boldsymbol{c} \in \mathcal{C}_{0:N}$.

At this point, we are ready to define a coherent risk measure. 
\vspace{0.2cm}
\begin{defn}[Coherent Risk Measure]\label{defi:coherent}{
We call the one-step conditional risk measures $\rho_t: \mathcal{C}_{t+1}\to \mathcal{C}_t$, $t=1,\ldots,N-1$ as in~\eqref{eq:dynriskmeasure} a \emph{coherent risk measure} if it satisfies the following conditions
\begin{itemize}
    \item \textbf{Convexity:} $\rho_t(\lambda c + (1-\lambda)c') \le \lambda \rho_t(c)+(1-\lambda)\rho_t(c')$, for all $\lambda \in (0,1)$ and all $c,c' \in \mathcal{C}_{t+1}$;
    \item \textbf{Monotonicity:} If $c\le c'$ then $\rho_t(c) \le \rho_t(c')$ for all $c,c' \in \mathcal{C}_{t+1}$;
    \item \textbf{Translational Invariance:} $\rho_t(c+c')=c+\rho_t(c')$ for all $c \in \mathcal{C}_t$ and $c' \in \mathcal{C}_{t+1}$;
    \item \textbf{Positive Homogeneity:} $\rho_t(\beta c)= \beta \rho_t(c)$ for all $c \in \mathcal{C}_{t+1}$ and $\beta \ge 0$.
\end{itemize}
}
\end{defn}
\vspace{0.2cm}

% Henceforth, all the risk measures considered are assumed to be coherent. An important example of a coherent risk measure is the  conditional value at risk (cVaR) given as
% \begin{equation}
%     \rho_t(Z_{t+1}) = \inf_{U \in \mathcal{C}_t} \left\{ U + \frac{1}{\alpha} \mathbb{E}\left[  (Z_{t+1}-U)_{+} \mid \mathcal{F}_t    \right]                        \right\},
% \end{equation}
% where $(\cdot)_{+}=\max\{\cdot, 0\}$ and the infimum should be understood point-wise. In general, the level $\alpha$ may be $\mathcal{F}_t$-measurable function with values in the interval $(0,1)$. In practice, we often assume $U \in \mathbb{R}$ and $\alpha \in (0,1)$.

Henceforth, all the risk measures considered are assumed to be coherent. In this paper, we are interested in the discounted infinite horizon problems. Let $\gamma \in (0,1)$ be a given discount factor. For $N=0,1,\ldots$, we define the functionals 
\begin{multline}
    \rho^\gamma_{0,t}(c_0,\ldots,c_t) = \rho_{0,t}(c_0,\gamma c_1,\ldots, \gamma^{t}c_t) \nonumber \\
                                      = \rho_0 \bigg(c_0 + \rho_{1} \big( \gamma c_{1}+\rho_{2}(\gamma^2c_{2}+\cdots \nonumber\\
                                      ~~+\rho_{t-1}\left(\gamma^{t-1}c_{t-1}+\rho_{N}(\gamma^{t}c_t) \right) \cdots )\big) \bigg),
\end{multline}
which are the same as~\eqref{eq:dynriskmeasure} for $t=0$, but with discounting $\gamma^{t}$ applied to each $c_t$. Finally, we have total discounted risk functional $\rho^{\gamma}:\mathcal{C}\to \mathbb{R}$ defined as \begin{equation}\label{eq:totaldiscrisk} \rho^{\gamma}(\boldsymbol{c}) = \lim_{t \to \infty} \rho^\gamma_{0,t}(c_0,\ldots,c_t).\end{equation} From~\cite[Theorem 3]{ruszczynski2010risk}, we have that $\rho^{\gamma}$ is convex, monotone, and positive homogeneous. 

%\section{Risk-Averse MDPs}

\section{Constrained Risk-Averse MDPs} \label{sec:mdps}

Notions of coherent risk and dynamic risk measures discussed in the previous section have been developed and applied in microeconomics and mathematical finance fields in the past two decades~\cite{vose2008risk}. Generally, risk-averse decision making is concerned with the behavior of agents, e.g. consumers and investors, who, when exposed to uncertainty, attempt to lower that uncertainty. The agents avoid situations with unknown payoffs, in favor of situations with payoffs that are more predictable, even if they are lower.

In a Markov decision making setting, the main idea in risk-averse control is to replace the conventional risk-neutral conditional expectation of the cumulative cost objectives with the more general coherent risk measures. In a path planning setting, we will show in our numerical experiments that considering coherent risk measures will lead to significantly more robustness to environment uncertainty.

 In addition to risk-aversity, an autonomous agent is often subject to constraints, e.g. fuel, communication, or energy budgets. These constraints can also represent mission objectives, e.g. explore an area or reach a goal.  

% Let $d^i: \mathcal{S} \times \mathcal{A} \to \mathbb{R}_{\ge 0}$ with $i=1,2,\ldots,n_c$, be a set of constraint cost functions. 

Consider a stationary controlled Markov process $\{s_t\}$, $t=0,1,\ldots$, wherein policies, transition probabilities, and cost functions do not depend explicitly on time. Each policy $\pi = \{\pi_t\}_{t=0}^\infty$ leads to  cost sequences $\boldsymbol{c}_t=c(s_t,\alpha_t)$, $t=0,1,\ldots$ and $\boldsymbol{d}_t^i=d^i(s_t,\alpha_t)$, $t=0,1,\ldots$, $i=1,2,\ldots,n_c$. We define the dynamic risk of evaluating the $\gamma$-discounted cost of a policy $\pi$ as
\begin{equation}\label{eq:objrisk}
    J_{\gamma}(\kappa_0,\pi) = \rho^{\gamma} \big( c(s_0,\alpha_0),c(s_1,\alpha_1),\ldots \big),
\end{equation}
 and the $\gamma$-discounted dynamic risk constraints of executing policy $\pi$ as
\begin{multline}\label{eq:constraint}
 D_{\gamma}^i(\kappa_0,\pi)=   \rho^{\gamma}\left(  d^i(s_0,\alpha_0),d^i(s_1,\alpha_1),\ldots \right) \le \beta^i, \\ i=1,2,\ldots,n_c,
\end{multline}
where $\rho^{\gamma}$ is defined in equation~\eqref{eq:totaldiscrisk}, $s_0 \sim \kappa_0$, and $\beta^i>0$, $i=1,2,\ldots,n_c$, are given constants.

In this work, we are interested in addressing the following problem:
\vspace{0.2cm}
\begin{problem}\textit{
For a given Markov decision process, a discount factor $\gamma \in (0,1)$, and a total risk  functional $J_{\gamma}(\kappa_0,\pi)$ as in equation~\eqref{eq:objrisk} and total cost constraints~\eqref{eq:constraint} with $\{\rho_t\}_{t=0}^\infty$ being coherent risk measures, compute 
\begin{align}
\pi^* \in &~\argmin_{\pi} ~~J_{\gamma}(\kappa_0,\pi) \nonumber \\ & \text{subject to} \quad \boldsymbol{D}_{\gamma}(\kappa_0,\pi) \preceq \boldsymbol{\beta}.
\end{align}
}
\end{problem}
\vspace{0.2cm}
We call a controlled Markov process with the ``nested'' objective~\eqref{eq:objrisk} and constraints~\eqref{eq:constraint} a  \emph{constrained risk-averse} MDP. It was  previously demonstrated in~\cite{chow2015risk,osogami2012robustness} that such coherent risk measure objectives can account for  modeling errors and parametric uncertainty in MDPs. One interpretation for Problem 1 is that we are interested in policies that minimize the incurred costs in the worst-case in a probabilistic sense and at the same time ensure that the system constraints, \textit{e.g.}, fuel constraints, are not violated even in the  worst-case probabilistic scenarios. 

Note that in Problem 1 both the objective function and the constraints are in general non-differentiable and non-convex in policy $\pi$ (with the exception of total expected cost as the coherent risk measure $\rho^\gamma$~\cite{altman1999constrained}). Therefore, finding optimal policies in general may be hopeless. Instead, we find sub-optimal polices by taking advantage of a Lagrangian formulation and then using an optimization form of Bellman's equations.

% Note that in the case where $\rho^\gamma$ is the total expected cost, we have a constrained Markov decision process~\cite{altman1999constrained}. Indeed, Problem 1 is a generalization of constrained Markov decision processes to coherent risk measures. Indeed, we show in Section~\ref{} that our approach converges to those of constrained Markov decision processes when $\rho^\gamma$ is the total expected cost.

% Note that since the constraints~\eqref{eq:constraint} are affine in the policy variable~$\pi$, there always exist a solution that solves the constrained optimization problem. However, the problem of minimizing a total cost function subject to constraints on total coherent risk is not well-posed in general, since total coherent risk is not necessarily a convex function of $\pi$ (\cite{chow2014algorithms} attempts to minimze total cost subject to CVaR constraints assuming an optimal policy exists).

Next, we show that the constrained risk-averse problem is equivalent to a non-constrained inf-sup risk-averse problem thanks to the Lagrangian method. 

\begin{proposition}
Let $J_\gamma(\kappa_0)$ be the  value of Problem 1 for a given initial distribution $\kappa_0$ and discount factor $\gamma$. Then, (i) the value function satisfies 
\begin{equation}\label{eq:Dds}
    J_\gamma(\kappa_0) = \inf_{\pi}\sup_{\boldsymbol{\lambda} \succeq \boldsymbol{0}} L_{\gamma}(\pi,{\boldsymbol{\lambda}}),
\end{equation}
where 
\begin{align}\label{eq:lagrangian}
    L_{\gamma}(\pi,\boldsymbol{\lambda}) = J_{\gamma}(\kappa_0,\pi)+ \langle \boldsymbol{\lambda},\left(\boldsymbol{D}_{\gamma}(\kappa_0,\pi)-\boldsymbol{\beta}\right) \rangle,
\end{align}
is the Lagrangian function.\\
(ii) Furthermore, a policy $\pi^*$ is optimal for~Problem 1, if and only if $J_\gamma(\kappa_0)=\sup_{\boldsymbol{\lambda} \succeq \boldsymbol{0}}~L_{\gamma}(\pi^*,\boldsymbol{\lambda})$.
\end{proposition}
\vspace{0.2cm}

At any time $t$, the value of $\rho_t$ is $\mathcal{F}_t$-measurable and is allowed to depend on the entire history of the process $\{s_0,s_1,\ldots\}$ and we cannot expect to obtain a Markov optimal policy~\cite{ott2010markov}. In order to obtain Markov policies, we need the following property~\cite{ruszczynski2010risk}. 
\vspace{0.2cm}

\begin{defn}\label{assum1}\textit{
 Let $m,n \in [1,\infty)$ such that $1/m+1/n=1$ and
$$
\mathcal{P} = \big\{p \in \mathcal{L}_n(\mathcal{S}, 2^\mathcal{S}, \mathbb{P}) \mid \sum_{s' \in \mathcal{S}} p(s') \mathbb{P}(s')=1,~p\ge 0 \big\}.
$$
A one-step conditional risk measure $\rho_t:\mathcal{C}_{t+1}\to \mathcal{C}_t$ is a Markov risk measure with respect to the controlled Markov process $\{s_t\}$, $t=0,1,\ldots$, if there exist a risk transition mapping $\sigma_t: \mathcal{L}_m(\mathcal{S}, 2^\mathcal{S}, \mathbb{P}) \times \mathcal{S} \times \mathcal{P} \to \mathbb{R}$ such that for all $v \in \mathcal{L}_m(\mathcal{S}, 2^\mathcal{S}, \mathbb{P})$ and $\alpha_t \in \pi(s_t)$, we have
\begin{equation}
    \rho_t(v(s_{t+1})) = \sigma_t(v(s_{t+1}),s_t,p(s_{t+1}|s_t,\alpha_t)),
\end{equation}
where $p:\mathcal{S}\times Act \to \mathcal{P}$ is called the controlled kernel.
}
\end{defn}
\vspace{0.2cm}

In fact, if $\rho_t$ is a coherent risk measure, $\sigma_t$ also satisfies the properties of a coherent risk measure (Definition 3). In this paper, since we are concerned with MDPs, the controlled kernel is simply the transition function $T$.
\vspace{0.2cm}
\begin{assum}\label{assum1}\textit{
The one-step coherent risk measure $\rho_t$ is a Markov risk measure.
}
\end{assum}
\vspace{0.2cm}
% Note that the Markov risk transition mapping depends on the function $\phi$, the states $s$, and probability vector $p(s,\alpha)$. The dot in $\phi(s_t,\alpha_t, \cdot)$ on the right hand side of~\eqref{eq:markovtrans} represents a dummy variable that is integrated/summed out with respect to the $s_t$-th row of the transition probability matrix $p(s_t,\alpha_t)$. 

The simplest case of the risk
transition mapping is in the
conditional expectation case
$\rho_t(v(s_{t+1})) =
\mathbb{E}\{v(s_{t+1}) \mid
s_t,\alpha_t\}$, \textit{i.e.}, 
\begin{multline}\label{eq:fdffd}
\sigma\left\{v(s_{t+1}),s_t,p(s_{t+1}|s_t,\alpha_t) \right\} =
\mathbb{E}\{ v(s_{t+1})\mid s_t,\alpha_t\} \\ =
\sum_{s_{t+1} \in \mathcal{S}} v(s_{t+1}) T(s_{t+1}\mid s_t,\alpha_t).
\end{multline}
Note that in the total discounted expectation case $\sigma$ is a linear function in $v$ rather than a convex function, which is the case for a general coherent risk measures. In the next result, we show that we can find a lower bound to the solution to Problem 1 via solving an optimization problem.

\vspace{0.2cm}
\begin{theorem}\textit{
Consider an MDP~$\mathcal{M}$   with the nested risk objective~\eqref{eq:objrisk},  constraints~\eqref{eq:constraint}, and discount factor $\gamma \in (0,1)$. Let Assumption~\ref{assum1} hold, let $\rho_t,~t=0,1,\ldots$ be  coherent risk measures as described in Definition~\ref{defi:coherent}, and suppose $c(\cdot,\cdot)$ and $d^i(\cdot,\cdot)$, $i=1,2,\ldots,n_c$, be non-negative and upper-bounded. Then, the solution $(\boldsymbol{V}^*_\gamma,\boldsymbol{\lambda}^*)$ to the following optimization problem (Bellman's equation)
\begin{align}\label{eq:valueiteration}
  & \sup_{\boldsymbol{V}_\gamma,\boldsymbol{\lambda} \succeq \boldsymbol{0}}~~\langle \boldsymbol{\kappa_0},\boldsymbol{V}_\gamma\rangle - \langle \boldsymbol{\lambda},\boldsymbol{\beta} \rangle \nonumber  \\
        &\text{subject to} \nonumber  \\
        &V_\gamma(s) \le c(s,\alpha) + \langle \boldsymbol{\lambda}, \boldsymbol{d}(s,\alpha)\rangle \nonumber \\ &\quad \quad \quad         +\gamma \sigma\{ {V}_\gamma(s'),s,p(s'|s,\alpha) \},~\forall s \in \mathcal{S},~\forall \alpha \in {Act,}
% V(s) =& \min_{\alpha \in Act} \sup_{\boldsymbol{\lambda} \succeq \boldsymbol{0}} \Big( c(s,\alpha) + \lambda {d}(s,a)
%      \\&\quad \quad \quad \quad+\gamma \sigma\{ V^*(s'),s,p(s'|s,\alpha) \}\Big),~~\forall s \in \mathcal{S},
    %   & \sup_{V,\boldsymbol{\lambda} \succeq \boldsymbol{0}}~~\langle \kappa_0,V \rangle - \langle \lambda,\beta \rangle \nonumber  \\
    %     &\text{subject to} \nonumber  \\
    %     &V(b(s)) \le c(b(s),\alpha) + \langle \boldsymbol{\lambda}, \boldsymbol{d}(b(s),a)\rangle \nonumber 
    %     \\&\quad \quad \quad \quad+\gamma \sigma\{ V^*(b'(s)),b(s),p(b'(s)|b(s),\alpha) \},
\end{align}
satisfies
\begin{equation} \label{eq:lowerboundrisk}
    J_\gamma(\kappa_0) \ge \langle \boldsymbol{\kappa_0},\boldsymbol{V}^*_\gamma \rangle-\langle\boldsymbol{\lambda}^*,\boldsymbol{\beta}\rangle.
\end{equation}
}
\end{theorem}
\vspace{0.2cm}

One interesting observation is that if the coherent risk measure $\rho^t$ is the total discounted expectation, then  Theorem 1 is consistent with the result by~\cite{altman1999constrained}. 
\vspace{.2cm}
\begin{corollary}
Let the assumptions of Theorem 1 hold and let $\rho_t(\cdot) = \mathbb{E}(\cdot |s_t,\alpha_t)$, $t=1,2,\ldots$. Then the solution $(\boldsymbol{V}^*_\gamma,\boldsymbol{\lambda}^*)$ to  optimization~\eqref{eq:valueiteration} satisfies
$$
    J_\gamma(\kappa_0) = \langle \boldsymbol{\kappa_0},\boldsymbol{V}^*_\gamma \rangle-\langle\boldsymbol{\lambda}^*,\boldsymbol{\beta}\rangle.
    $$
    Furthermore, with $\rho_t(\cdot) = \mathbb{E}(\cdot |s_t,\alpha_t)$, $t=1,2,\ldots$, optimization~\eqref{eq:valueiteration} becomes a linear program.
\end{corollary}
\vspace{0.2cm}

% \begin{remark} \textit{
% In the conventional case of total expected costs and constraints, \textit{i.e.,}  $\rho^\gamma_{\kappa_0}(c_t) = \sum_t E_{\kappa_0}^\pi \gamma^t c_t$, we have $$\sigma(\phi,s,p(s,a)) = \sum_{s' \in \mathcal{S}}\phi(s,a,s')T(s'|s,a).$$ Then, we can show that 
% \begin{equation}\label{eq:exactbound}
% J_\gamma(\kappa_0) = \langle \kappa_0,V^*\rangle-\langle\lambda^*,\beta\rangle.
% \end{equation}
% % That is, solving convex optimization problem~\eqref{eq:valueiteration} gives the exact value of of the constrained problem. This result is consistent with previous work by~\cite{altman1999constrained} for total expected costs and constraints. Indeed,~\eqref{eq:exactbound} can also be inferred from the fact that, in the $\rho^\gamma_{\kappa_0}(c_t) = \sum_t E_{\kappa_0}^\pi \gamma^t c_t$ case, both the objective function and the constraints are linear in the policy $\pi$ and the fact that $\sum_t E_{\kappa_0}^\pi \gamma^t c_t+ \sum_t E_{\kappa_0}^\pi \gamma^t \langle \lambda, d_t \rangle=\sum_t E_{\kappa_0}^\pi \gamma^t (c_t+\langle \lambda, d_t \rangle)$.}
% % \end{remark}
% % \vspace{0.2cm}

Once the values of $\boldsymbol{\lambda}^*$ and $\boldsymbol{V}^*_\gamma$ are found by solving optimization problem~\eqref{eq:valueiteration}, we can find the policy as
\begin{align}
    \pi^*(s) \in &~\argmin_{\alpha \in Act}~\Big(  c(s,\alpha) + \langle \boldsymbol{\lambda}^*, \boldsymbol{d}(s,\alpha)\rangle        \nonumber \\ & \quad \quad \quad \quad +\gamma \sigma\{ V^*_\gamma(s'),s,p(s'|s,\alpha) \}   \Big).
\end{align}

Note that $\pi^*$ is a deterministic, stationary policy. Such policies are desirable in practical applications, since they are more convenient to implement on actual robots. Given an uncertain environment, $\pi^*$ can be designed offline and used for path planning. 

In the next section, we discuss methods to find solutions to optimization problem~\eqref{eq:valueiteration}, when $\rho^\gamma$ is an arbitrary coherent risk measure.

\section{DCPs for Constrained Risk-Averse MDPs}

Note that since $\rho^\gamma$ is a coherent, Markov risk measure (Assumption 1), $v \mapsto \sigma(v,\cdot,\cdot)$ is convex (because $\sigma$ is also a coherent risk measure). Next, we demonstrate that optimization problem~\eqref{eq:valueiteration} is indeed a DCP. Re-formulating equation~\eqref{eq:valueiteration} as a minimization yields
\begin{align}\label{eq:valueiteration2}
  & \inf_{\boldsymbol{V}_\gamma,\boldsymbol{\lambda} \succeq \boldsymbol{0}}~~ \langle \boldsymbol{\lambda},\boldsymbol{\beta} \rangle - \langle \boldsymbol{\kappa_0},\boldsymbol{V}_\gamma\rangle \nonumber  \\
        &\text{subject to} \nonumber  \\
        &V_\gamma(s) \le c(s,\alpha) + \langle \boldsymbol{\lambda}, \boldsymbol{d}(s,\alpha)\rangle \nonumber \\ &\quad \quad \quad         +\gamma \sigma\{ {V}_\gamma(s'),s,p(s'|s,\alpha) \},~\forall s \in \mathcal{S},~\forall \alpha \in {Act}.
% V(s) =& \min_{\alpha \in Act} \sup_{\boldsymbol{\lambda} \succeq \boldsymbol{0}} \Big( c(s,\alpha) + \lambda {d}(s,a)
%      \\&\quad \quad \quad \quad+\gamma \sigma\{ V^*(s'),s,p(s'|s,\alpha) \}\Big),~~\forall s \in \mathcal{S},
    %   & \sup_{V,\boldsymbol{\lambda} \succeq \boldsymbol{0}}~~\langle \kappa_0,V \rangle - \langle \lambda,\beta \rangle \nonumber  \\
    %     &\text{subject to} \nonumber  \\
    %     &V(b(s)) \le c(b(s),\alpha) + \langle \boldsymbol{\lambda}, \boldsymbol{d}(b(s),a)\rangle \nonumber 
    %     \\&\quad \quad \quad \quad+\gamma \sigma\{ V^*(b'(s)),b(s),p(b'(s)|b(s),\alpha) \},
\end{align}

At this point, we define $f_0(\boldsymbol{\lambda})=\langle \boldsymbol{\lambda},\boldsymbol{\beta} \rangle$, $g_0(\boldsymbol{V}_\gamma)=\langle \boldsymbol{\kappa_0},\boldsymbol{V}_\gamma\rangle$, $f_1({V}_\gamma)={V}_\gamma$, $g_1(\boldsymbol{\lambda})=c + \langle \boldsymbol{\lambda}, \boldsymbol{d}\rangle$, and $g_2({V}_\gamma)=\gamma \sigma(V_\gamma,\cdot,\cdot)$. Note that $f_0$ and $g_1$ are convex (linear) functions of $\boldsymbol{\lambda}$ and $g_0$, $f_1$, and $g_2$ are convex functions in $\boldsymbol{V}_\gamma$. Then, we can re-write~\eqref{eq:valueiteration2} as
\begin{align}\label{eq:DCP}
      & \inf_{\boldsymbol{V}_\gamma,\boldsymbol{\lambda} \succeq \boldsymbol{0}}~~ f_0(\boldsymbol{\lambda})-g_0(\boldsymbol{V}_\gamma)  \nonumber  \\
        &\text{subject to} \nonumber  \\
        &f_1({V}_\gamma)-g_1(\boldsymbol{\lambda})-g_2({V}_\gamma) \le 0, ~~\forall s,\alpha.
\end{align}

In fact, optimization problem~\eqref{eq:DCP} is a standard DCP~\cite{horst1999dc}. DCPs arise in  many applications, such as feature selection in  machine learning~\cite{le2008dc} and inverse covariance estimation in statistics~\cite{thai2014inverse}. Although DCPs can be solved globally~\cite{horst1999dc}, \textit{e.g.} using branch and bound algorithms~\cite{lawler1966branch}, a locally optimal solution can be obtained based on techniques of nonlinear optimization~\cite{Bertsekas99} more efficiently. In particular, in this work, we use a variant of the convex-concave procedure~\cite{lipp2016variations,shen2016disciplined}, wherein  the concave terms are replaced by a convex upper bound and solved. In fact, the disciplined convex-concave programming (DCCP)~\cite{shen2016disciplined} technique linearizes DCP problems into a (disciplined) convex program (carried out automatically via the DCCP Python package~\cite{shen2016disciplined}), which is then converted into an equivalent cone program by
replacing each function with its graph implementation. Then, the cone program can be solved readily by available convex programming solvers, such as CVXPY~\cite{diamond2016cvxpy}. 

At this point, we should point out that solving~\eqref{eq:valueiteration} via the 
DCCP method, finds the (local) saddle points to optimization problem ~\eqref{eq:valueiteration}. However, from Theorem 1, we have that every saddle point to~\eqref{eq:valueiteration} satisfies~\eqref{eq:lowerboundrisk}. In other words, every saddle point corresponds to a lower bound to the optimal value of Problem~1. 

\subsection{DCPs for CVaR and EVaR Risk Measures}

In this section, we present the specific DCPs for finding the risk value functions for two coherent risk measures studied in our numerical experiments, namely, CVaR and EVaR. 

For a given confidence level $\varepsilon \in (0,1)$, value-at-risk ($\mathrm{VaR}_\varepsilon$) denotes the $(1-\varepsilon)$-quantile value of the cost variable. $\mathrm{CVaR}_\varepsilon$ measures the expected loss in the $(1-\varepsilon)$-tail given that the particular threshold $\mathrm{VaR}_\varepsilon$ has been crossed. $\mathrm{CVaR}_\varepsilon$ is given by 
\begin{equation}
    \rho_t(c_{t+1}) = \inf_{\zeta \in \mathbb{R}} \left\{ \zeta + \frac{1}{\varepsilon} \mathbb{E}\left[  (c_{t+1}-\zeta)_{+} \mid \mathcal{F}_t    \right]                        \right\},
\end{equation}
where $(\cdot)_{+}=\max\{\cdot, 0\}$. A value of $\varepsilon \simeq 1$ corresponds to a risk-neutral policy; whereas, a value of $\varepsilon \to 0$ is rather a risk-averse policy. 

In fact, Theorem 1 can applied to CVaR since it is a coherent risk measure. For an MDP $\mathcal{M}$, the risk value functions can be computed by DCP~\eqref{eq:DCP}, where 
$$
g_2({V}_\gamma) = \inf_{\zeta\in \mathbb{R}} \left\{ \zeta + \frac{1}{\varepsilon} \sum_{s' \in \mathcal{S}}\left(V_\gamma(s')-\zeta\right)_{+} T(s'\mid s,\alpha)        \right\},
$$
where the infimum on the right hand side of the above equation can be absorbed into the overal infimum problem, \textit{i.e.,} $\inf_{\boldsymbol{V}_\gamma,\boldsymbol{\lambda} \succeq \boldsymbol{0},\zeta}$. Note that $g_2({V}_\gamma)$ above is  convex in $\zeta$~\cite[Theorem 1]{rockafellar2000optimization} because the function $(\cdot)_+$ is increasing and convex~\cite[Lemma A.1., p. 117]{ott2010markov}.

Unfortunately, CVaR ignores the losses below the VaR threshold. EVaR is the tightest upper bound in the sense of Chernoff inequality for the value at risk (VaR) and CVaR and its dual representation is associated with the relative entropy. In fact, it was shown in~\cite{ahmadi2017analytical} that $\mathrm{EVaR}_\varepsilon$ and $\mathrm{CVaR}_\varepsilon$ are equal only if there are no losses ($c\to -\infty$) below the $\mathrm{VaR}_\varepsilon$ threshold. In addition, EVaR is a strictly monotone risk measure; whereas, CVaR is only monotone~\cite{ahmadi2019portfolio}. $\mathrm{EVaR}_\varepsilon$  is given by
\begin{equation}
    \rho_t(c_{t+1}) = \inf_{\zeta >0} \left(  {\log \left(\frac{\mathbb{E}[e^{\zeta c_{t+1}} \mid \mathcal{F}_t]}{\varepsilon}\right)/ \zeta}        \right).            
\end{equation}
Similar to $\mathrm{CVaR}_\varepsilon$, for $\mathrm{EVaR}_\varepsilon$, $\varepsilon \to 1$ corresponds to a risk-neutral case; whereas, $\varepsilon\to 0$ corresponds to a risk-averse case. In fact, it was demonstrated in~\cite[Proposition 3.2]{ahmadi2012entropic} that $\lim_{\varepsilon\to 0} \mathrm{EVaR}_{\varepsilon}(Z) = \esssup(Z)$. 

 Since $\mathrm{EVaR}_\varepsilon$ is a coherent risk measure, the conditions of Theorem 1 hold. Since $\zeta>0$, using the change of variables, $\tilde{\boldsymbol{V}}_\gamma \equiv \zeta {\boldsymbol{V}}_\gamma$ and $\tilde{\boldsymbol{\lambda}} \equiv \zeta {\boldsymbol{\lambda}}$ (note that this change of variables is monotone increasing in $\zeta$~\cite{agrawal2018rewriting}), we can compute EVaR value functions by solving~\eqref{eq:DCP}, where
 $$
 \begin{cases}
 f_0(\tilde{\boldsymbol{\lambda}})=\langle \tilde{\boldsymbol{\lambda}},\boldsymbol{\beta} \rangle,\\
 f_1(\tilde{V}_\gamma)=\tilde{V}_\gamma,\\
 g_0(\tilde{\boldsymbol{V}}_\gamma)=\langle \boldsymbol{\kappa_0},\tilde{\boldsymbol{V}}_\gamma\rangle,\\
 g_1(\tilde{\boldsymbol{\lambda}})=\zeta c + \langle \tilde{\boldsymbol{\lambda}}, \boldsymbol{d}\rangle,~\text{and} \\
 g_2(\tilde{{V}_\gamma}) =   \log\left(\frac{\sum_{s' \in \mathcal{S}}e^{ \tilde{V}_\gamma(s' )}T(s'|s,\alpha)}{\varepsilon}\right).
 \end{cases}
 $$
 
Similar to the CVaR case, the infimum over $\zeta$ can be lumped into  the overall infimum problem, \textit{i.e.,} $\inf_{\tilde{\boldsymbol{V}}_\gamma,\tilde{\boldsymbol{\lambda}} \succeq \boldsymbol{0},\zeta>0}$. Note that $g_2(\tilde{\boldsymbol{V}}_\gamma)$ is  convex in  $\tilde{V}_\gamma$, since the logarithm of sums of exponentials is convex~\cite[p.~72]{boyd2004convex}. The lower bound in~\eqref{eq:lowerboundrisk} can then be obtained as $\frac{1}{\zeta}\left(\langle \boldsymbol{\kappa_0},\tilde{\boldsymbol{V}}_\gamma\rangle-\langle \tilde{\boldsymbol{\lambda}},\boldsymbol{\beta} \rangle\right)$.

\section{Related Work and Discussion}

We believe this work is the first study of constrained MDPs with both coherent risk objectives and constraints. We emphasize that our method leads to a  policy that lower-bounds the value of Problem 1 for general coherent, Markov risk measures. In the case of no constraints $\boldsymbol{\lambda}\equiv 0$, our proposed method can also be applied to risk-averse MDPs with no constraints.

With respect to risk-averse MDPs, \cite{tamar2016sequential,tamar2015policy} proposed a sampling-based algorithm for finding saddle point solutions to MDPs with total coherent risk measure costs using policy gradient methods. \cite{tamar2016sequential} relies on the assumption that the risk envelope appearing in the dual representation of the coherent risk measure is known with an explicit canonical convex programming formulation. As the authors indicated, this is the case for CVaR, mean-semi-deviation, and spectral risk measures~\cite{shapiro2014lectures}. However, such explicit form is not known for  general coherent risk measures, such as EVaR. Furthermore, it is not clear whether the saddle point solutions are a lower bound or upper bound to the optimal value. Also, policy-gradient based methods require calculating the gradient of  the coherent risk measure, which is not available in explicit form in general. MDPs with CVaR constraint and total expected costs were studied in~\cite{prashanth2014policy,chow2014algorithms} and locally optimal solutions were found via policy gradients, as well. However, this method also leads to saddle point solutions and cannot be applied to general coherent risk measures. In addition, since the objective and the constraints are described by different coherent risk measures, the authors assume there exists a policy that satisfies the CVaR constraint (feasibility assumption), which may not be the case in general. 

Following the footsteps of~\cite{pflug2016time}, a promising approach based on approximate value iteration was proposed for MDPs with CVaR objectives in~\cite{chow2015risk}. But, it is not clear how one can extend this method to other coherent risk measures. An infinite-dimensional linear program was derived analytically for MDPs with coherent risk objectives  in~\cite{haskell2015convex} with implications for solving chance and stochastic-dominance constrained MDPs. Successive finite approximation of such infinite dimensional linear programs was suggested leading to sub-optimal solutions. Yet, the method cannot be extended to constrained MDPs with coherent risk objectives and constraints. A policy iteration algorithm for finding policies that minimize total coherent risk measures for MDPs was studied in~\cite{ruszczynski2010risk,fan2018process} and a computational non-smooth Newton method was proposed in~\cite{ruszczynski2010risk}. Our work, extends~\cite{ruszczynski2010risk,fan2018process} to constrained problems and uses a DCCP computational method, which takes advantage of already available software (DCCP and CVXPY).

\vspace{-0.2cm}
\section{Numerical Experiments}\label{sec:example}

In this section, we evaluate the proposed methodology with a numerical example. In addition to the traditional total expectation, we consider two other coherent risk measures, namely, CVaR and EVaR. All experiments were carried out using a MacBook Pro with 2.8 GHz Quad-Core Intel Core i5 and 16 GB of RAM. The resultant linear programs and DCPs were solved using CVXPY~\cite{diamond2016cvxpy} with DCCP~\cite{shen2016disciplined} add-on in Python.

\subsection{Rover MDP Example Set Up}

 \begin{figure}[t] \label{fig:gwa_again}\centering{
\includegraphics[scale=.3]{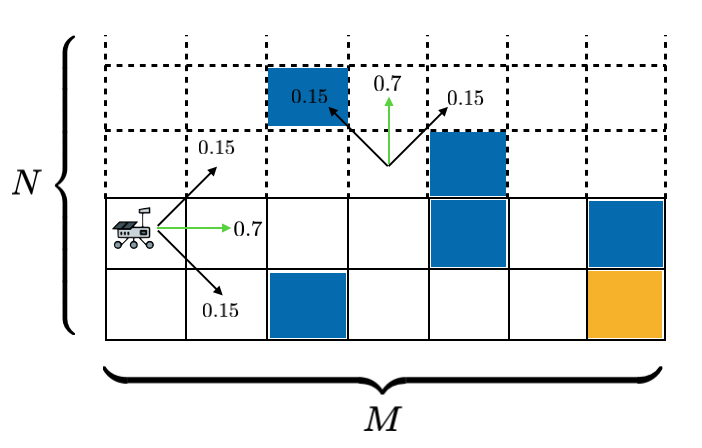}
\vspace{-0.4cm}
\caption{Grid world illustration for the rover navigation example. Blue cells denote the obstacles and the yellow cell denotes the goal.}
\vspace{-.5cm}
} 
 \end{figure}

An  agent (e.g. a rover) has to autonomously navigate a two dimensional terrain map (e.g. Mars surface) represented by an $M \times N$ grid. A rover can move from cell to
cell. Thus, the state space is given by
 $ \mathcal{S} = \{s_{i}|i=x+My,x\in\{1,\dots,M\},y \in \{1,\dots,N\}\}. $ 
The action set available to the robot is $ Act = \{E,\ W,\ N,\ S,\ NE,\ NW,\ SE,\ SW\}$, \textit{i.e.,} diagonal moves are allowed. 
The actions move the robot from its current cell to a neighboring
cell, with some uncertainty. The state transition probabilities for various cell types are shown for actions $E$ and $N$ in Figure 2.  Other actions lead to analogous transitions.  

% In the POMDP setting, partial observability arises because the rover cannot determine obstacle cell
% location from measurements directly. The observation space is
% $ \mathcal{O}=\{o_{i} |i=x+My, x\in\{1,\dots,M\},y\in\{1,\dots,N\}\}.$ Once at an adjacent cell to an obstacle, the rover can identify an actual obstacle position (dark green) with probability $0.6$, and a distribution over the nearby cells (light green). 

% {\color{blue} STILL IN PROGRESS! ONE EXAMPLE WITH CVAR AND ONE WITH EVAR}

% % This example is adapted from~\cite{chow2015risk}, where we added partial observability to the autonomous agent.
% An  agent (e.g. a robot) has to autonomously navigate a two dimensional terrain map (e.g. Mars surface) represented by a $10\times 10$ grid world ($100$ states) with $15$ obstacles of different shapes. At each time step the agent can move to any of its eight neighboring states (diagonal moves are allowed). Due to sensing and control noise, however, with probability $\delta$ a move to a random neighboring state occurs.
With regards to constraint costs, there is an stage-wise cost of
each move until reaching the goal of $2$, to account for fuel usage constraints. In between the
starting point and the destination, there are a number of obstacles that the agent should
avoid. Hitting an obstacle incurs the immediate cost  of $10$, while the goal grid region has zero immediate cost. These two latter immediate costs are captured by the cost function. The discount factor is $\gamma=0.95$. 

% After a move is chosen, the observation of  the agent is assumed to be binary, \textit{i.e.}, either an obstacle is detected in the next cell that the robot is moving to or not. 
 
The objective is to compute a safe path that is fuel efficient, \textit{i.e.,} solving Problem 1. To this end, we consider total expectation, CVaR, and EVaR as the coherent risk measure. 

As a robustness test, inspired by~\cite{chow2015risk}, we included a set of single grid obstacles that are perturbed in a random direction to one of the neighboring grid
 cells with probability $0.2$ to represent uncertainty in the terrain map. For each risk measure, we run $100$ Monte Carlo simulations with the calculated policies and count failure rates, \textit{i.e.,} the number of times a collision has occurred during a run.

% CVaR is given by 
% \begin{equation}
%     \rho_t(c_{t+1}) = \inf_{z \in \mathbb{R}} \left\{ z + \frac{1}{\alpha} \mathbb{E}\left[  (c_{t+1}-z)_{+} \mid \mathcal{F}_t    \right]                        \right\},
% \end{equation}
% where $(\cdot)_{+}=\max\{\cdot, 0\}$ and the infimum should be understood point-wise. In general, the confidence level $\alpha$ may be $\mathcal{F}_t$-measurable function with values in the interval $(0,1)$. Here, we assume  $\alpha \in (0,1)$. A value of $\alpha \simeq 1$ corresponds to a risk-neutral policy; whereas, a value of $\alpha \simeq 0$ is rather a risk-averse policy. For CVaR risk measure, \eqref{eq:valueiterationsfc} can be computed as
% \begin{align*}
% \hspace{-.5cm}    V_{\gamma,\mathcal{M}}([s,g]) =& \sum_{\alpha , g' , o }   \omega(g',\alpha \mid g,o) O(o|g') c([s,g],\alpha) \\&+ \gamma\inf_{z \in \mathbb{R}} \bigg\{ z + \frac{1}{\alpha} \sum_{g',s', o,\alpha} \left( V\left([s',g']\right)-z\right)_{+} \\
%     &~~\times O(o\mid s)\omega(g',\alpha \mid g, o) T(s'\mid s,\alpha)\bigg\},
% \end{align*}
% where the infimum on the right hand side of the above equation can either be solved by line search techniques or
% by representation in terms of an elementary linear programming problem since it is convex in $z$~\cite[Theorem 1]{rockafellar2000optimization} (the function $(\cdot)_+$ is increasing and convex~\cite[Lemma A.1., p. 117]{ott2010markov}).

\subsection{ Results}

%  \begin{figure*}[!h] \label{fig:mdp}\centering{
% %\includegraphics[scale=.3]{mdpexample.png}\\
% \includegraphics[scale=.23]{mdpnorisk.eps}
% \includegraphics[scale=.24]{mdpcvar2.eps}
% \includegraphics[scale=.24]{mdpcvar.eps}
% \vspace{-0.5cm}
% \caption{Results for the MDP example with total expectation (left), CVaR (middle), and EVaR (right) coherent risk measures. The goal is located at the yellow cell. Notice the 10 single cell obstacles used for robustness test.}
% %\vspace{-0.5cm}
% } 
%  \end{figure*}

In our experiments, we consider three grid-world sizes of $10\times 10$, $15 \times 15$, and $20 \times 20$ corresponding to $100$, $225$, and $400$ states, respectively. For each grid-world, we allocate 25\% of the grid to obstacles, including 3, 6, and 9 uncertain (single-cell) obstacles for the $10\times 10$, $15 \times 15$, and $20 \times 20$ grids, respectively.  In each case, we solve DCP~\eqref{eq:valueiteration} (linear program in the case of total expectation) with $|\mathcal{S}||Act|=MN \times 8 = 8MN$ constraints and $MN+2$ variables (the risk value functions $V_\gamma$'s, Langrangian coefficient $\lambda$, and $\zeta$ for CVaR and EVaR). In these experiments, we set $\varepsilon= 0.15$ for CVaR and EVaR coherent risk measures. The fuel budget (constraint bound $\beta$) was set to 50, 10, and 200 for the $10\times 10$, $15 \times 15$, and $20 \times 20$ grid-worlds, respectively. The initial condition was chosen as $\kappa_0(s_M)=1$, \textit{i.e.,} the agent starts at the right most grid at the bottom.

A summary of our numerical experiments is provided in Table 1. Note the computed values of Problem 1 satisfy $\mathbb{E}(c)\le \mathrm{CVaR}_\varepsilon(c) \le \mathrm{EVaR}_\varepsilon(c)$. This is in accordance with the theory that EVaR is a more conservative coherent risk measure than CVaR~\cite{ahmadi2012entropic}. 

\begin{table}[t!]
\label{table:comparison}
\centering
\setlength\tabcolsep{2.5pt}
\begin{tabular}{lccccc} \midrule
\makecell{$(M \times N)_{\rho_t}$}  & \makecell{$J_\gamma(\kappa_0)$}  & \makecell{ Total \\  Time [s] }  & \makecell{\# U.O.}   & \makecell{ F.R.}   \\ 
\midrule
$(10\times 10)_{\mathbb{E}}$           & 5.10                          & 0.7 & 3                & 9\%                                        \\[1.5pt]
$(15\times 15)_{\mathbb{E}}$          & 7.53                          &  1.0  & 6           & 18\%                                             \\[1.5pt]
$(20\times 20)_{\mathbb{E}}$         & 8.98                          &  1.6 & 9               & 21\%                                         \\[1.5pt]
\midrule
$(10\times 10)_{\text{CVaR}_{\varepsilon}}$            & $\ge$7.76                          &  5.4 & 3            &1\%                                            \\[1.5pt]
$(15\times 15)_{\text{CVaR}_{\varepsilon}}$           & $\ge$9.22                          & 8.3   & 6           &3\%                                          \\[1.5pt]
$(20\times 20)_{\text{CVaR}_{\varepsilon}}$         & $\ge$12.76                          & 10.5   & 9          &5\%                                           \\[1.5pt]
\midrule
$(10\times 10)_{\text{EVaR}_{\varepsilon}}$            & $\ge$7.99                         & 3.2  & 3          &0\%                                            \\[1.5pt]
$(15\times 15)_{\text{EVaR}_{\varepsilon}}$          & $\ge$11.04                         & 4.9         &  6         &0\%                                      \\[1.5pt]
$(20\times 20)_{\text{EVaR}_{\varepsilon}}$            & $\ge$15.28                         & 6.6      & 9           &2\%                                       \\[1.5pt]
\midrule
\end{tabular}
\caption{Comparison between total expectation, CVaR, and EVaR coherent risk measures. $(M \times N)_{\rho_t}$ denotes experiments with grid-world of size $M \times N$ and one-step coherent risk measure $\rho_t$. $J_\gamma(\kappa_0)$ is the valued of the constrained risk-averse problem (Problem 1). Total Time denotes the time taken by the CVXPY solver to solve the associated linear programs or DCPs in seconds. $\#$ U.O. denotes the number of single grid uncertain obstacles used for robustness test.  F.R. denotes the failure rate out of 100 Monte Carlo simulations with the computed policy. }
\end{table}

  \begin{figure*}[!h] \label{fig:mdp}\centering{
\includegraphics[scale=.25]{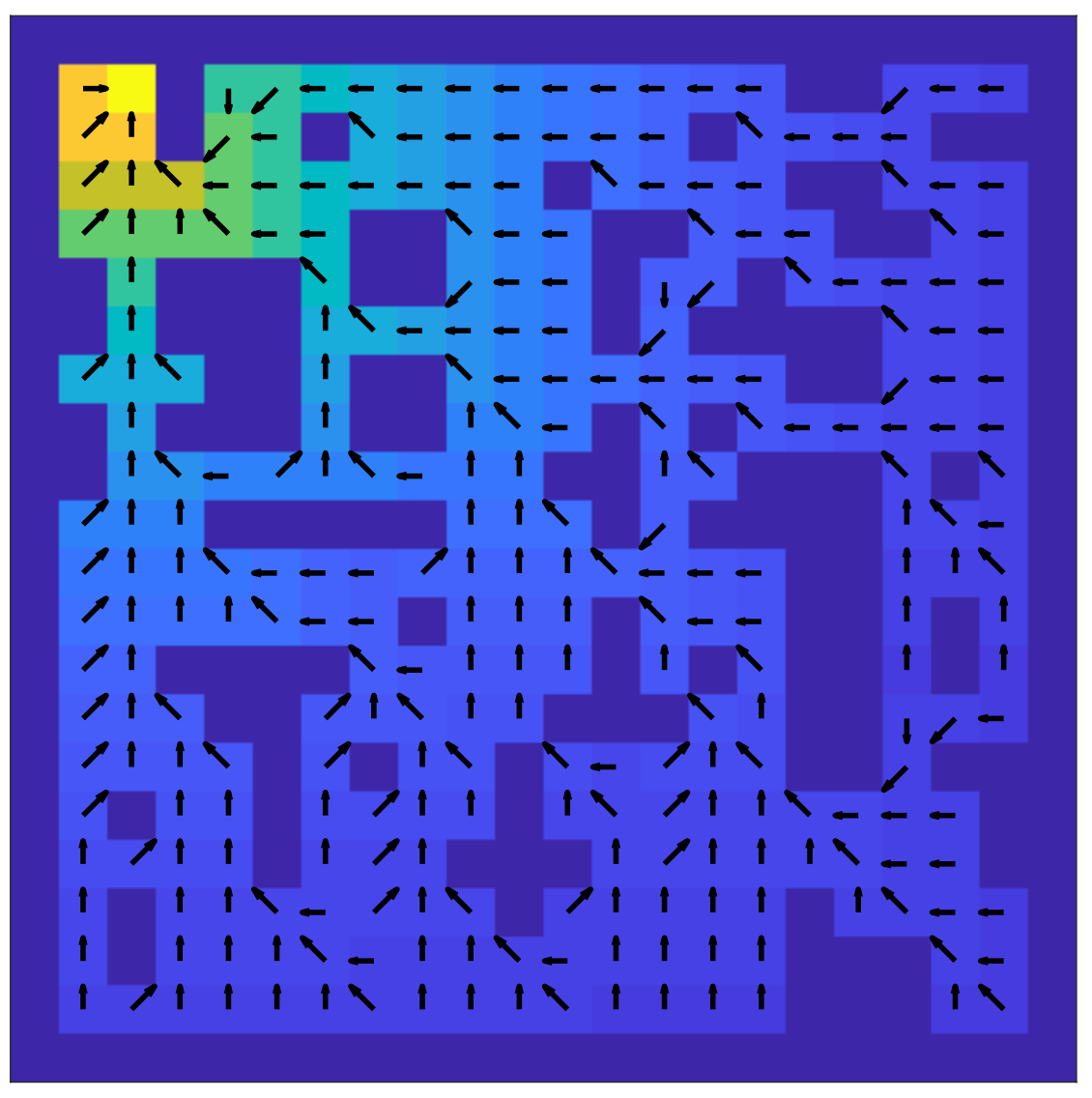}~
\hspace{1cm}
\includegraphics[scale=.25]{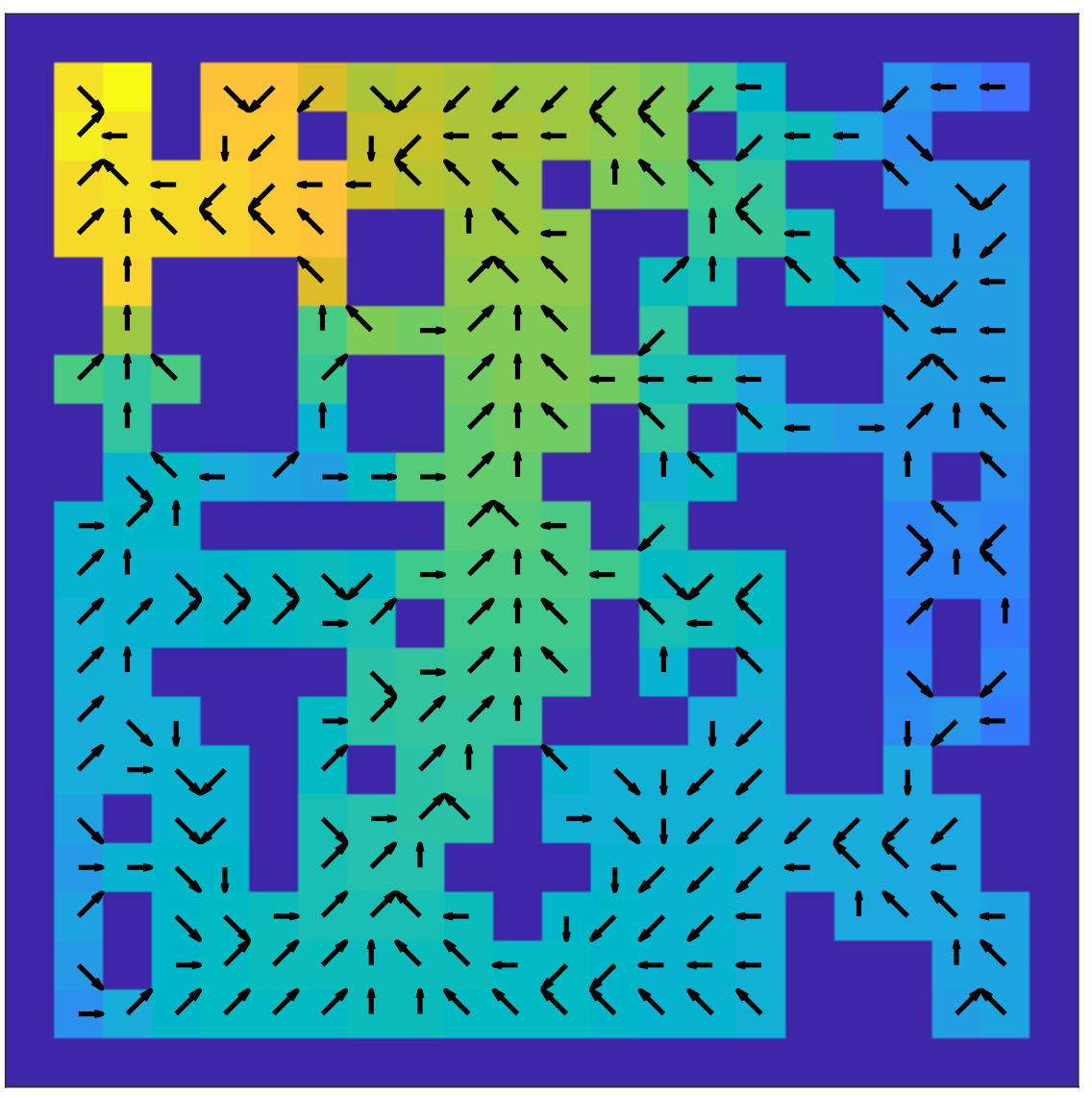}
\hspace{1cm}
\includegraphics[scale=.25]{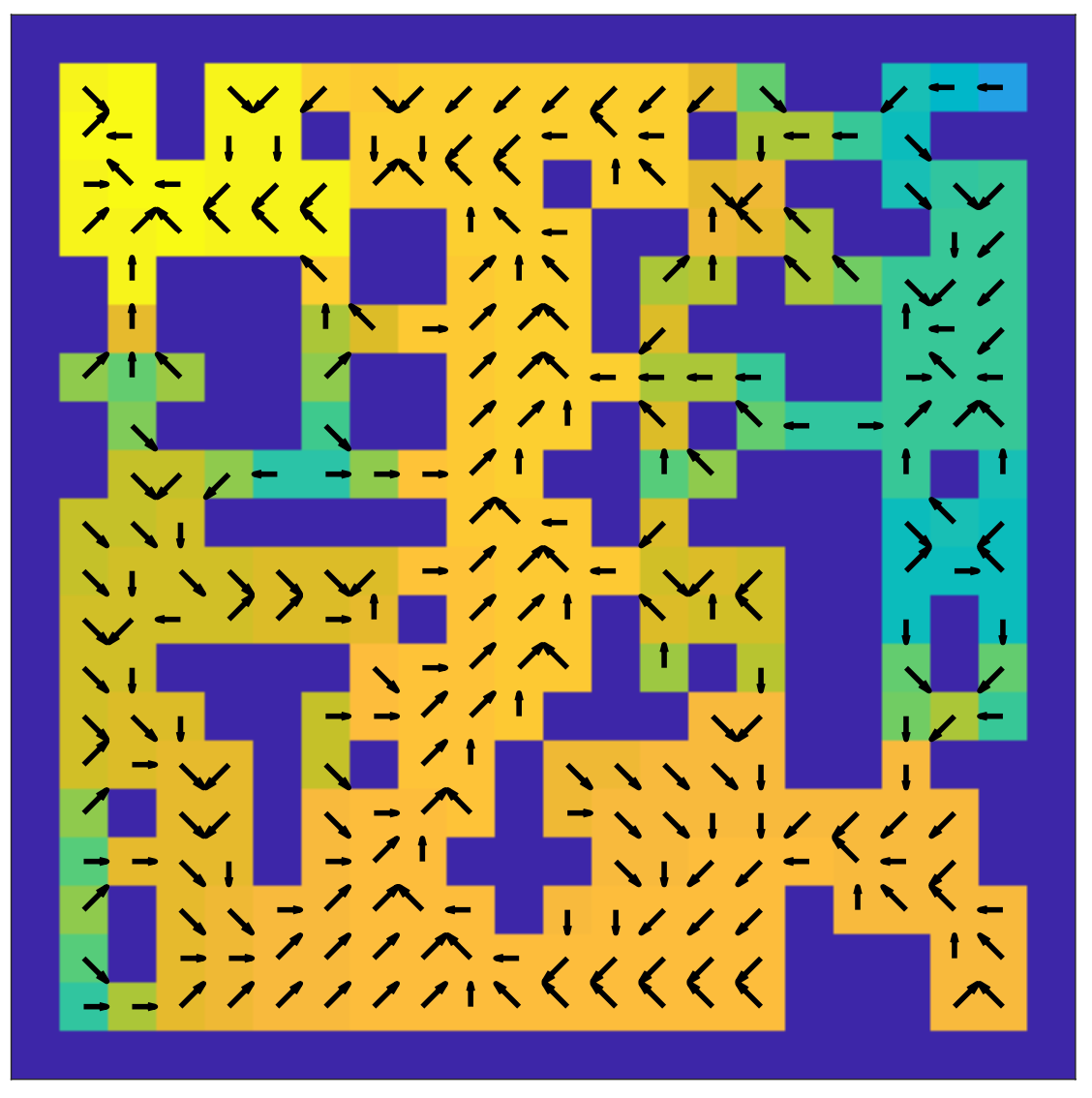}
\vspace{-0cm}
\caption{Results for the MDP example with total expectation (left), CVaR (middle), and EVaR (right) coherent risk measures. The goal is located at the yellow cell. Notice the 9 single cell obstacles used for robustness test.}
%\vspace{-0.5cm}
} 
 \end{figure*}

For total expectation coherent risk measure, the calculations took significantly less time, since they are the result of solving a set of linear programs. For CVaR and EVaR, a set of DCPs were solved. CVaR calculation was the most computationally involved. This observation is consistent with~\cite{ahmadi2019portfolio} were it was discussed that EVaR calculation is much more efficient than CVaR. Note that these calculations can be carried out offline for policy synthesis and then the policy can be applied for risk-averse robot path planning.

The table also outlines the failure ratios of each risk measure. In this case, EVaR outperformed both CVaR and total expectation in terms of robustness, tallying with the fact that EVaR is conservative. In addition, these results suggest that, although discounted total expectation is a measure of performance in high number of Monte Carlo simulations, it may not be practical to use it for real-world planning under uncertainty scenarios. CVaR and especially EVaR seem to be a more reliable metric for performance in planning under uncertainty.

For the sake of illustrating the computed policies, Figure~3 depicts the results obtained from solving DCP~\eqref{eq:valueiteration} for a $20\times 20$ grid-world. The arrows on grids depict the (sub)optimal actions and the heat map indicates the values of Problem 1 for each grid state.  Note that the values for EVaR are greater than those for CVaR and the values for CVaR are greater from those of total expectation. This is in accordance with the theory that $\mathbb{E}(c)\le \mathrm{CVaR}_\varepsilon(c) \le \mathrm{EVaR}_\varepsilon(c)$~\cite{ahmadi2012entropic}. In addition, by inspecting the computed actions in obstacle dense areas of the grid-world (for example, the top right area), we infer that the actions in more risk-averse cases (especially, EVaR) have a higher tendency to steer the robot away from the obstacles given the diagonal transition uncertainty as depicted in Figure 2; whereas, for total expectation, the actions are rather concerned about taking the robot to the goal region.
\vspace{-0.3cm}

\section{Conclusions} \label{sec:conclusions}

We proposed an optimization-based method for finding sub-optimal policies that lower-bound the constrained risk-averse problem in MDPs. We showed that this optimization problem is in DCP form for general coherent risk measures and can be solved using DCCP method. Our methodology generalized constrained MDPs with total expectation risk measure to general coherent, Markov risk measures. Numerical experiments were provided to show the efficacy of our approach.

In this paper, we assumed the states are fully observable. In future work, we will consider extension of the proposed framework to Markov processes with partially observable states~\cite{ahmadi2020risk,fan2015dynamic,fan2018risk}. Moreover, we only considered discounted infinite horizon problems in this work. We will explore risk-averse policy synthesis in the presence of other cost criteria~\cite{carpin2016risk,cavus2014risk} in our prospective research. In particular, we are interested in risk-averse planning in the presence of high-level mission specifications in terms of linear temporal logic formulas~\cite{wongpiromsarn2012receding,ahmadi2020stochastic}.

\bibliographystyle{aaai21}
\bibliography{references}

\clearpage

\end{document}